\documentclass[sigconf, nonacm, screen]{acmart}

\AtBeginDocument{%
  }

\setcopyright{acmlicensed}
\copyrightyear{2025}
\acmYear{2025}
\acmDOI{XXXXXXX.XXXXXXX}

\acmConference[ICCPS]{International Conference on Cyber-Physical Systems}{May 06--09,
  2025}{Irvine, CA}
\acmISBN{978-1-4503-XXXX-X/18/06}

\usepackage{algorithm}
\usepackage{algpseudocode}
\usepackage{graphicx}
\usepackage{subcaption}

\begin{document}

\title[RLS3: RL-Based Synthetic Sample Selection]{RLS3: RL-Based Synthetic Sample Selection to Enhance \\ Spatial Reasoning in Vision-Language Models for \\ Indoor Autonomous Perception}


\author{Joshua R. Waite}
\affiliation{%
  \institution{Iowa State University}
  \city{Ames}
  \state{IA}
  \country{USA}}
\email{jrwaite@iastate.edu}

\author{Md. Zahid Hasan}
\affiliation{%
  \institution{Iowa State University}
  \city{Ames}
  \state{IA}
  \country{USA}}
\email{zahid@iastate.edu}

\author{Qisai Liu}
\affiliation{%
  \institution{Iowa State University}
  \city{Ames}
  \state{IA}
  \country{USA}}
\email{supersai@iastate.edu}

\author{Zhanhong Jiang}
\affiliation{%
  \institution{Iowa State University}
  \city{Ames}
  \state{IA}
  \country{USA}}
\email{zhjiang@iastate.edu}

\author{Chinmay Hegde}
\affiliation{
\institution{New York University}
  \city{New York}
  \state{NY}
  \country{USA}}
\email{chinmay.h@nyu.edu}

\author{Soumik Sarkar}
\affiliation{%
  \institution{Iowa State University}
  \city{Ames}
  \state{IA}
  \country{USA}}
\email{soumiks@iastate.edu}







\renewcommand{\shortauthors}{Waite et al.}

\begin{abstract}
  Vision-language model (VLM) fine-tuning for application-specific visual grounding based on natural language instructions has become one of the most popular approaches for learning-enabled autonomous systems. However, such fine-tuning relies heavily on high-quality datasets to achieve successful performance in various downstream tasks. Additionally, VLMs often encounter limitations due to insufficient and imbalanced fine-tuning data. To address these issues, we propose a new generalizable framework to improve VLM fine-tuning by integrating it with a reinforcement learning (RL) agent. Our method utilizes the RL agent to manipulate objects within an indoor setting to create synthetic data for fine-tuning to address certain vulnerabilities of the VLM. Specifically, we use the performance of the VLM to provide feedback to the RL agent to generate informative data that efficiently fine-tune the VLM over the targeted task (e.g. spatial reasoning). The key contribution of this work is developing a framework where the RL agent serves as an informative data sampling tool and assists the VLM in order to enhance performance and address task-specific vulnerabilities. By targeting the data sampling process to address the weaknesses of the VLM, we can effectively train a more context-aware model. In addition, generating synthetic data allows us to have precise control over each scene and generate granular ground truth captions. Our results show that the proposed data generation approach improves the spatial reasoning performance of VLMs, which demonstrates the benefits of using RL-guided data generation in vision-language tasks.
\end{abstract}



\begin{teaserfigure}
\centering
  \includegraphics[width=0.9\textwidth]{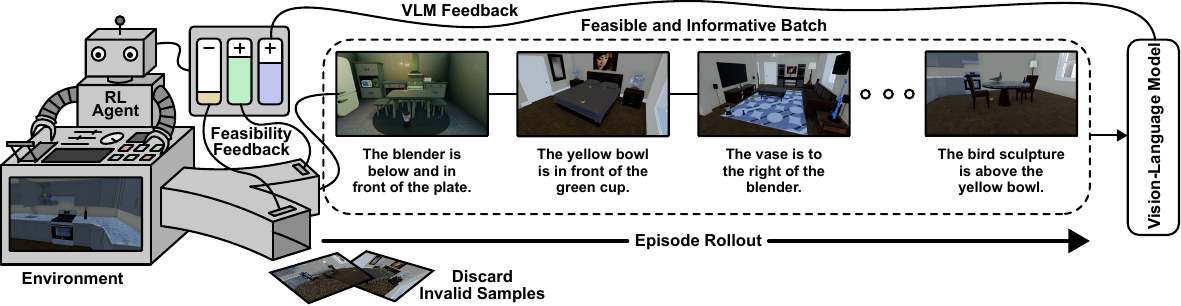}
  \caption{Overview of RLS3 to generate a feasible and informative batch for fine-tuning a VLM to improve spatial reasoning.}
  \Description{A graphical overview of RLS3. An RL agent is shown interacting with an environment to generate samples. The RL agent receives feasibility feedback from the environment and VLM feedback from the VLM. A few example images with captions are shown, such as "The blender is below and in front of the plate."}
  \label{fig:teaser}
\end{teaserfigure}


\maketitle

\section{Introduction}

Autonomous perception has progressed significantly with the advent of deep learning, enabling its deployment in a wide range of cyber-physical systems applications. Vision-language models (VLMs) especially have proven to be powerful tools due to enhanced understanding of real-world scenes, which is critical for precisely interacting with environments in robotics~\cite{gao2024physicallygroundedvisionlanguagemodels}. This characteristic also makes VLMs well-suited for distracted driving detection~\cite{10492662} and fully autonomous self-driving cars~\cite{tian2024drivevlm, 10531702}.  However, many challenges remain, and different models have various strengths and weaknesses.

In particular, CLIP-type models, which perform contrastive vision-language fine-tuning, have shown that natural language model fine-tuning can lead to very high performance on a particular scenario utilizing user-specific datasets to pair visual and textual representations effectively. 
Although existing CLIP-type models have achieved high performance, they heavily rely on the quality of pretraining datasets, assuming that the image-text pairs are of high quality and in perfect one-to-one correspondence. This dependency often leads to challenges since the available datasets can be insufficient or imbalanced. Such limitations can stop the models' ability to generalize effectively, particularly in scenarios involving rare or complex visual-linguistic relationships. Another relevant, recently released model, PaliGemma~\cite{beyer2024paligemmaversatile3bvlm}, addresses some of these generalization issues by leveraging transfer learning capabilities, allowing it to adapt to specialized tasks. However, this comes at the cost of reduced ``out-of-the-box" performance.

One area where VLMs have particular difficulty is spatial reasoning. This task is especially important in autonomous perception when similar objects need to be distinguished from one another by their relative positions, e.g., ``the leftmost bottle" or ``The bottle to the right of the bowl." One potential cause for this is that VLMs may consider the input as a bag of concepts rather than maintaining the order of the words and how spatial terms relate to them~\cite{yuksekgonul2023when}. Another consideration is that there are typically very limited samples with explicit spatial context in vision-language datasets.  


However, some recent work has tried to address this by curating data with explicit spatial context~\cite{cheng2024spatialrgpt, Chen_2024_CVPR}. Such data curation processes often struggle to generate reliable labels for data without ground truth, leading to accumulation of training data noise. It significantly affects the performance of supervised fine-tuning strategies and large-scale datasets are typically needed to alleviate this problem to some extent. 
Another way to address these limitations in model fine-tuning is through the use of reinforcement learning (RL), which offers a more adaptive and efficient approach. RL from human feedback (RLHF)~\cite{NIPS2017_d5e2c0ad,stiennon2022learningsummarizehumanfeedback,NEURIPS2022_b1efde53} has emerged as an effective strategy for aligning models, especially large language models (LLMs), with human preferences. 
Reinforced Fine-Tuning (ReFT) offers a framework aimed at enhancing reasoning capabilities through RL-based fine-tuning~\cite{luong2024reftreasoningreinforcedfinetuning}. By applying RL in this context, models can continuously adapt and refine their outputs based on how well they meet specific reasoning tasks. 
Active learning~\cite{ren2021surveydeepactivelearning} is another group of strategies in which training data is selectively sampled from a larger dataset to maximize performance gain. However, despite the progress made with active learning, RLHF, and frameworks like ReFT, these approaches are limited by their dependence on existing labeled datasets, which can hinder their ability to generalize to unseen or complex tasks. 
Synthetic data offers a promising solution by generating large volumes of precisely labeled data without the labor-intensive process of manual annotation~\cite{johnson2017clevr}. This enables the models to learn from a broader, more diverse dataset, which can be tailored to include rare or challenging scenarios that might be underrepresented in real-world data. Additionally, synthetic data has proven effective in training models for deployment in a variety of real-world applications~\cite{waite2023activeshooterdetectionrobust,9308468, Martinez-Gonzalez2020, Tremblay_2018_CVPR_Workshops}. 




To address these challenges, we propose a new framework that leverages a reinforcement learning agent to dynamically generate targeted synthetic data for fine-tuning VLMs, for which a high-level graphical overview is shown in Fig.~\ref{fig:teaser}. Rather than relying solely on static datasets, our approach enables the RL agent to manipulate high-fidelity simulators to create challenging scenarios, particularly those in which the VLMs exhibit weaknesses, such as spatial reasoning. By providing feedback directly from the VLM, our method positions the RL agent as a feasible and informative data sampler, focusing on manipulating object positions within a scene to create scenarios where the VLM demonstrates poor spatial reasoning performance. This approach involves using spatial positioning sentences generated from the ground truth and pairing them with image data generated by the RL agent. 
By fine-tuning the VLM with this targeted synthetic data, we can address specific vulnerabilities, particularly in spatial reasoning.
This method allows for the evaluation and improvement of spatial reasoning in popular VLMs, which makes sure that they become more robust and effective in handling diverse vision-language relationships.
Our experimental results show that this method improves the VLM's ability to handle complex spatial relationships, demonstrating the potential of RL agents as powerful tools for data augmentation and model fine-tuning. The implications of this work extend beyond VLM, suggesting a general framework for improving AI models through intelligent data generation.


Our main contributions are summarized as follows: (i) We present a generalizable framework that enables integrated data sampling for targeted model fine-tuning across diverse applications. (ii) We introduce an innovative approach that integrates reinforcement learning (RL) agents with vision-language models (VLM) to dynamically generate feasible and informative data, specifically addressing the VLM's weaknesses and enhancing its generalization capabilities. (iii) Our method provides continuous feedback from the VLM to the RL agent, ensuring sustained improvements and robustness over time. (iv) Extensive experiments demonstrate the effectiveness of our approach, showing significant performance improvements in tasks involving complex spatial reasoning. 


\section{Related Work}
\label{sec:relatedworks}



\subsection{Enhancing VLM Spatial Reasoning} 
The development of vision-language models has led to significant advancements in artificial intelligence, particularly in tasks that require a joint understanding of both visual and textual information. However, most general-purpose VLMs like CLIP ~\cite{radford2021learning}
and BLIP ~\cite{li2022blip} show limitations when it comes to spatial reasoning. Specifically, spatial reasoning tasks require a deeper understanding of the surrounding environment, such as determining relative positions, spatial relationships, and contextual object arrangements within images.

There have been notable works attempting to enhance the spatial reasoning capabilities of VLMs. SpatialVLM~\cite{Chen_2024_CVPR} introduces a 3D Visual Question Answering (VQA) data generation framework, which improves spatial reasoning by employing three-dimensional data. However, the framework requires a significant amount of data to be effective because of the noisy labels. Another recent effort, SpatialRGPT~\cite{cheng2024spatialrgpt}, introduced a data curation pipeline and the integration of depth data into the vision encoder of the VLM. However, one limitation of SpatialRGPT is the complexity of the depth data integration, which can lead to higher computational costs and a need for specialized hardware. Grounding DINO~\cite{liu2023grounding} offers another perspective on enhancing spatial reasoning in VLMs by focusing on object grounding. They enable the identification and spatial localization of objects by explicitly linking them in the visual domain to their corresponding textual descriptions. However, the model is limited by its reliance on well-annotated data, which can constrain its generalizability in environments with limited labeled samples.
\subsection{RL in Model Fine-Tuning}
The success of LLMs has been partially attributed to RL from human feedback (RLHF) that was introduced by Christiano et al.\cite{NIPS2017_d5e2c0ad} for training RL agents using human preferences, which led to significant improvements in task performance. 
In addition, Ouyang et al.\cite{NEURIPS2022_b1efde53} presented InstructGPT, which uses human feedback to better follow user instructions. Furthermore, Bai et al.~\cite{bai2022training} further explored using RLHF to align language models with desirable traits such as helpfulness and honesty, which is crucial in high-stakes decision-making.
The ReFT framework~\cite{luong2024reftreasoningreinforcedfinetuning} integrates RL with LLMs to enhance reasoning in VLMs, particularly in complex reasoning tasks. Similarly, Wu et al.~\cite{wu2023fine} introduced a fine-grained RL approach for optimizing reasoning capabilities in LLMs, which shows improvements in decision-making accuracy. 
Recent works have also explored the integration of RL with foundational models to enhance their capabilities. For instance, Ye et al. ~\cite{pushingrlboundaries2023} investigate how LLMs and VLMs can be incorporated into RL frameworks to improve learning efficiency and task performance. Another work by Zhai et al.~\cite{zhai2024finetuninglargevisionlanguagemodels} proposed treating VLMs as decision-making agents by fine-tuning them with RL. In their approach, the VLM processes environment observations and pre-designed prompts as input and then outputs actions represented in textual form along with chain-of-thought reasoning and using generated actions to interact with the environment. 


\subsection{Synthetic Data Generation}
Data generation remains a critical strategy for improving the performance of VLMs. Traditional methods, such as image transformations and synthetic data generation, have been enhanced by advanced techniques. Neural Inverse Rendering ~\cite{sengupta2023neural}, for instance, generates realistic training data by reconstructing detailed 3D scenes from single images, thus enriching the variety and quality of training datasets. Universal Simulator (UniSim) ~\cite{unisim2023} provides a versatile environment for creating diverse datasets, enabling the simulation of complex real-world interactions and scenarios. Re-Thinking Inverse Graphics with Large Language Models ~\cite{kulits2024rethinkinginversegraphicslarge}
explores the use of LLMs for improving inverse graphics techniques, offering new insights into model interpretability and functionality. 



\subsection{Active Learning}

Active learning is a widely used technique that seeks to improve model performance by selectively querying the most informative samples from a pool of data. One popular approach is uncertainty sampling, which prioritizes data points where the model exhibits the highest uncertainty~\cite{10.5555/3305381.3305504, doi:10.1177/14759217221150376, Yoo_2019_CVPR, houlsby2011bayesianactivelearningclassification}. Core-set selection seeks to select a subset of data points that best represent the full dataset, improving model generalization by ensuring diversity in the selected examples\cite{10.1145/3534678.3539476}. BatchBALD~\cite{houlsby2011bayesianactivelearningclassification} and BADGE~\cite{DBLP:conf/iclr/AshZK0A20} both aim to utilize the core principles of these strategies to select data that maximize information gain while maintaining sample diversity. While these methods select samples from an existing dataset, our approach introduces a novel distinction by generating new samples. By using RL agents to generate synthetic data, we can create tailored examples specifically designed to address model weaknesses, particularly in complex scenarios that would not be well-represented in existing datasets. A recent work, GENESIS-RL~\cite{yang2024genesis}, presented a similar approach that leverages an RL agent to generate potentially unsafe edge cases for autonomous systems. 










\section{Preliminaries}
In this section, we present some background knowledge of RL and VLM fine-tuning as preliminaries to characterize the methodology introduced in the next section. We start with the RL in the sequel.

\begin{figure*}[t]
\centering
\includegraphics[width=0.98\linewidth,trim={0 2mm 0 0},clip]{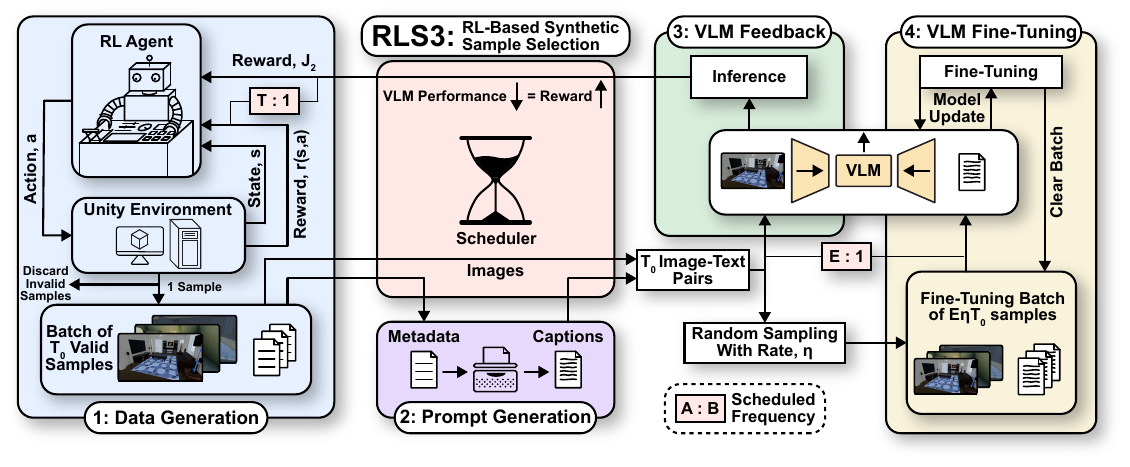}
\caption{A detailed overview of our proposed RLS3 framework. The scheduler acts as a synchronizer between the processes. The loop begins with data generation where the RL agent takes $T$ timesteps in which $T_0 \leq T$ image-metadata samples are generated from the Unity environment. In each step, the agent receives an intrinsic reward based on the feasibility of the generated sample. The episode is paused on the last step to allow the following processes to complete. Next, the metadata is used in the prompt generation process to create captions describing a spatial relation. The $T_0$ image-text pairs are then inputted to the VLM for inference. The performance of the VLM is used as an extrinsic reward signal $J_2$ for the RL agent at the end of the episode. Steps 1, 2, and 3 repeat for $E$ RL episodes to generate a diverse batch of data for fine-tuning. Diversity is further increased in the fine-tuning batch by sampling the generated data with sampling rate $\eta$ for each episode, resulting in a fine-tuning batch size of $E \eta T_0$. After fine-tuning the VLM finished, the batch is cleared and the process repeats for the next iteration.}
\Description{A detailed overview of the RLS3 framework. The framework is shown broken into 4 main sections: 1. Data Generation, 2. Prompt Generation, 3. VLM Feedback, and 4. VLM Fine-Tuning. Simples icons are added to each step described in the caption.}
\label{fig:spatialCLIP}
\end{figure*}

\subsection{Markov Decision Process}
In this context, we use the standard RL problem formulation as a Markov Decision Process (MDP)~\cite{agarwal2019reinforcement,kaelbling1996reinforcement}. We denote by $\mathcal{M}=\{\mathcal{S},\mathcal{A},p,r,\gamma\}$ an MDP, where $\mathcal{S}$ signifies the state space, $\mathcal{A}$ indicates the action space, $p(s'|s,a):\mathcal{S}\times \mathcal{A}\to\mathcal{S}$ denotes the transition dynamics, $r(s,a):\mathcal{S}\times \mathcal{A}\to \mathbb{R}$ represents the reward function and $\gamma\in [0,1]$ is the discount factor. The goal of an MDP is to learn a policy denoted by $\pi(a|s):\mathcal{S}\to\mathcal{A}$ such that it can maximize the total discounted return $J(\pi)=\mathbb{E}_{\pi}\bigg[\sum_{t=0}^T\gamma^tr(s_t,a_t)\bigg]$, where $T$ is the episodic length. $\pi(a|s)\in[0,1]$ intuitively denotes probability of the policy choosing the action $a$ at the given state $s$. Note that the RL agent in this study behaves like a generator to generate valid data samples that can be used to test the VLM component in the proposed pipeline, which to some extent deviates from its regular applications in critical decision-making processes. More details will be disclosed in the next section. 

\subsection{VLM Fine-Tuning}
VLMs typically take paired vision and language data samples as input for fine-tuning. To this end, VLMs comprise a text encoder $\phi$ and an image encoder $\psi$, which correspondingly map the text and image inputs into a joint feature space. Therefore, training a VLM leads to the alignment of text and image modalities by maximizing their feature similarities.   
Suppose that we have a mini-batch of $N$ image-text pairs $\{x_i,y_i\}_{i=1}^N$ sampled from a large training dataset during each training iteration. The image encoder first encodes $x_i$ into image features $z_i$, i.e., $z_i=\psi(x_i)\in\mathbb{R}^d$. The next step is to obtain the text features $w_i, i\in\{1,2,...,N\}$ by using the text encoder $\phi$. Then we have $w_i=\phi(y_i)\in\mathbb{R}^d$.
Thereby, the VLM loss can be written as follows:
\begin{equation}\label{eq_unify}
    \mathcal{L}:=(\mathcal{L}_{I\to T}+\mathcal{L}_{T\to I})/2,
\end{equation}
where $
\mathcal{L}_{I\to T} := -\frac{1}{N}\sum_{i=1}^N\textnormal{log}\frac{\textnormal{exp}(f(z_i, w_i; \tau))}{\sum_{j=1}^N\textnormal{exp}(f(z_i, w_j;\tau))},
$ and $
\mathcal{L}_{T\to I} := -\frac{1}{N}\sum_{i=1}^N\textnormal{log}\frac{\textnormal{exp}(f(z_i, w_i; \tau))}{\sum_{j=1}^N\textnormal{exp}(f(z_j, w_i; \tau))},
$
$f(\cdot, \cdot;\cdot)$ is the alignment function for image and text, which is typically the cosine similarity. $\tau$ is the temperature parameter for the softmax function. Particularly, $f(z_i, w_i;\tau) = \frac{z_i\cdot w_i}{\tau\|z_i\|_2\|w_i\|_2}$, where $\|\cdot\|_2$ is the Euclidean norm. 
In our study, we use two different VLM model architectures, CLIP~\cite{radford2021learning} (and its variants, e.g., NegCLIP~\cite{yuksekgonul2023when} 
as a baseline), and PaliGemma~\cite{beyer2024paligemmaversatile3bvlm} to facilitate the understanding of their spatial reasoning capabilities. 
However, their objective functions are practically the same as described in Eq.~\ref{eq_unify}, while they differ slightly in model architectures.  
\section{Methods}
\label{sec:method}


In this section, we introduce our proposed framework, RLS3, which is depicted in Fig.~\ref{fig:spatialCLIP}. Each component is the framework is presented in detail below. 

\subsection{Unity Environment}
\begin{figure}[ht]
\centering
\includegraphics[width=0.85\linewidth]{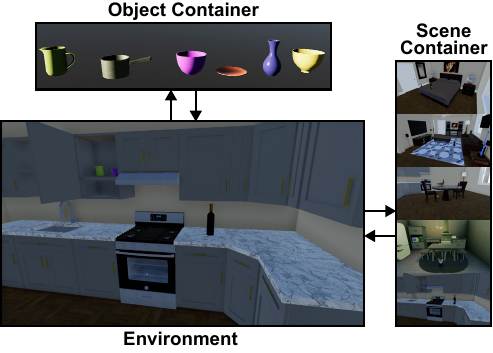}
\caption{A diagram showing the Unity environment structure. There is one active scene at a time, which is cycled over the episode. Each step selects one of the three active objects and is swapped with another in the object container. Only a handful of the available objects are shown here.}
\Description{}
\label{fig:unity_env}
\end{figure}

We utilize the Unity~\cite{unity} game engine environment in this paper to generate synthetic data (scene images). An important consideration when working with synthetic data is the feasibility of the generated samples. As such, we utilize the in-built constraints within Unity to generate images of feasible scenarios with a focus on spatial relations. The environment is constructed using assets available on the Unity Asset Store and provides several semi-realistic scenes of indoor settings, including rooms such as a kitchen, bedroom, and living room. We have multiple cameras and objects that can be moved to facilitate greater variety in the generated data. Objects to move are selected three at a time from a list of nine available objects. Each of the five scenes has a camera and designated surfaces on which the chosen objects can be placed. Along with guaranteeing that the objects are placed on the surfaces, there is also a check for whether the moved object would overlap with another object, ensuring only feasible scenarios are selected for fine-tuning. The ground truth data generated includes the object names, coordinates, and rotation, as well as the camera coordinates and rotation. An overview of the unity environment can be seen in Fig.~\ref{fig:unity_env}.

We use the Unity ML-Agent package~\cite{juliani2020} to facilitate communication between Unity and Python. However, we do not use the integrated RL agent and instead use a package called Stable Baselines 3~\cite{stable-baselines3}, which will be discussed more in the following section. 

\subsection{Data Generation With an RL Agent}
As shown in Fig.~\ref{fig:spatialCLIP}, when Unity interacts with the RL agent, it is used as an environment to provide state observation $s$ and the intrinsic reward $r(s,a)$, after the RL agent executes an action $a$. In this context, the state $s\in\mathbb{R}^n$ includes the following: the index of the object being moved, scene index, size and center of the surfaces in the scene (such as a tabletop), coordinates and rotation of all objects, and camera coordinates and rotation. Note that for all the objects, we only take the coordinates of the central location into consideration such that each of them can be regarded as a 3D point. In this study, the state space is discrete and $n=32$. Also, there are always three objects in each scene, though the proposed framework is applicable to scenes with more than three objects. Action $a\in\mathbb{R}^3$ by the RL agent is the movement of only an object in each time step, which can be quantified through a difference between the coordinates of the object before and after the movement. The intrinsic reward $r$ is calculated based on the validity of the movement. For example, if the movement of the object results in unrealistic scenes as objects overlapping or out of the table bounds, which implies invalid coordinates, then $r=-1$. Otherwise, all valid movements lead to $r=1$. We denote by $\mathcal{A}_{va}$ and $\mathcal{A}_{in}$ the valid and invalid action spaces.
Specifically, at a given state $s$, it can be described as
\begin{equation}
    r(s,a)=\begin{cases}
        1 & a\in\mathcal{A}_{va} \\
        -1 & a\in\mathcal{A}_{in}
    \end{cases}
\end{equation}
As the RL agent is trained to obtain the optimal policy that can execute actions to push the Unity to generate valid images, valid movements are critical. One may wonder how many valid images are required for the subsequent VLM fine-tuning. It can be regarded as a key hyperparameter to tune in practice. Moreover, it has an impact on the episodic length $T$. To ensure that there are sufficient valid images involving valid movements for fine-tuning the VLM, we can set the number of images as the threshold for $T$ such that $T\geq T_0$, where $T_0$ is the bare minimum number of valid images during one episode. Based on Fig.~\ref{fig:spatialCLIP}, it is immediately known that $T_0=N$. The entire generated valid image set is denoted by $\{x_i\}_{i=1}^N$. Additionally, each image is saved with a ground truth description. For the RL agent training, we select the soft actor-critic (SAC)~\cite{haarnoja2018soft}, which is a popular off-policy algorithm widely adopted in different areas. After the interactions between RL agent and Unity over a trajectory, the cumulative discounted reward is calculated, denoted by $J_1$. However, the updates of the actor and critic networks are not ready as the RL agent requires the VLM performance signal to augment the reward such that the generated valid images $\{x_i\}_{i=1}^N$ are sent to a prompt generator and then a VLM for inference. To ensure feasible samples are prioritized, we initially pre-train the RL agent for $100k$ timesteps with only the intrinsic reward and without interacting with the other processes of the RLS3 framework.





\subsection{Angle-Based Prompt Generation}

\begin{figure}[ht]
\centering
\includegraphics[width=0.9\linewidth]{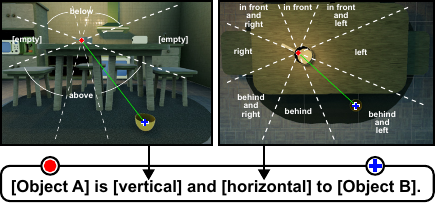}
\caption{Illustration of our template-based prompt generation. The center of `Object A' is located on the origin of the eight regions in the horizontal direction and 3 regions in the vertical direction in which spatial terms are selected. The horizontal regions are aligned such that `behind' is facing towards the camera. The cameras are aligned with the horizontal and vertical axes in this figure to more clearly show the regions. The generated caption for this scenario is: ``The small pot is above, behind and to the left of the yellow bowl."}
\Description{}
\label{fig:angle-prompt}
\end{figure}

Angle-based prompt provides a specific spatial positional relation of an object relative to a target object. Denote by $g(\cdot,\cdot)$ the angle-based prompt generator. First, we randomly select an object $o^1_i$ present in the scene $x_i$ as a primary object. Next, another object $o^2_i$ is picked as the secondary object and the prompt generator creates a prompt describing the spatial relation between these two objects. Mathematically, it is expressed by $y_i=g(o^1_i,o^2_i)$, where $y_i$ is the textual description for each $x_i$ based on the prompt generator. Similarly, we denote by $\{y_i\}_{i=1}^N$ the text set for all images. As shown in Fig.~\ref{fig:angle-prompt}, the surrounding space of a primary object (small pan) is divided into eight horizontal 45-degree regions with their corresponding spatial terms. Then the position of the center of a secondary object is measured to determine the region where it is located. Then, one or two spatial words are attached while creating the prompt. As a result, the prompt captures the horizontal positional context. A similar approach is used to capture the vertical positional relation where if the measured angle between object centers falls within a $\pm$ 20-degree region, only the horizontal relation is used. Otherwise, a vertical relation is also appended. If the angle between the object centers is steeper than $\pm$ 75 degrees, then only the vertical relation is used in the generated prompt. 
These conditions allow the creation of prompts with varying levels of complexity due to the range of one to three spatial terms used.

To shed light on the spatial reasoning capabilities of VLMs, we resort to two categories of models for VLM fine-tuning to examine their performance, consisting of PaliGemma and CLIP. For CLIP, another model, NegCLIP, is also taken into account as a baseline. NegCLIP conducts \textit{data augmentation} to the original dataset $\{x_i,y_i\}_{i=1}^N$. Particularly, ``negative samples" are introduced to boost the model performance and robustness, which is made by deliberately modifying the generated text $y_i$ associated with the image $x_i$. The modification can be with a spatial term swapped or with the object order swapped in the text. Hence, for CLIP-type VLM fine-tuning, it will have at most totally $3N$ samples, denoted by $\{x_i,(y_i, \tilde{y}_i, \hat{y}_i)\}_{i=1}^N$, where $\tilde{y}_i$ signifies the one with a spatial term swapped in the text, while $\hat{y}_i$ the object order swapped. The motivation for the data augmentation is that vanilla CLIP evidently performs poorly~\cite{yuksekgonul2023when}. For PaliGemma, we use the question-answer format with the ``positive caption" as the answer with questions of the form, ``What is the position of [Object A] relative to [Object B]?"



\subsection{VLM Evaluation and Fine-Tuning}
After receiving the generated dataset of $N$ image-text pairs $\{x_i,y_i\}_{i=1}^N$ or the augmented ones $\{x_i,(y_i, \tilde{y}_i, \hat{y}_i)\}_{i=1}^N$, we are now ready to evaluate the VLM with inference. For PaliGemma, we compute a rubric-based score $S_{PG}$ to quantify spatial reasoning capabilities. The rubric is as follows:
\begin{itemize}
    \item \textbf{Score = 5}: All spatial terms are correct.
    \item \textbf{Score = 4}: 2 of 3 spatial terms are correct.
    \item \textbf{Score = 3}: 1 of 2 spatial terms is correct.
    \item \textbf{Score = 2}: 1 of 3 spatial terms is correct.
    \item \textbf{Score = 1}: No spatial terms are correct.
    \item \textbf{Penalty -1 (minimum score of 1)}: Opposite spatial term is used (e.g., `right' instead of `left').
    \item \textbf{Penalty -1 (minimum score of 1)}: Too many spatial terms are used (e.g., 'to the right and behind' instead of 'behind').
\end{itemize}
Rather than directly using $S_{PG}$ as a reward signal, we invert it such that a low score corresponds to a higher reward.
For CLIP, Eq~\ref{eq_unify} is adopted with a minor modification by changing $N$ to $3N$. In practice, the total number of data samples can range from $2N$ to $3N$, depending on how data augmentation is performed. However, this will not change the loss calculation. So far, we have obtained the loss for the VLM fine-tuning, which is the augmented reward signal for the original RL training. Analytically, it can be represented by
\begin{equation}
    J_2=\begin{cases}
        \mathcal{L}^2 & \textnormal{with CLIP}\\
        (6-S_{PG})^2 & \textnormal{with PaliGemma}\\
    \end{cases}
\end{equation}
Thereby, the augmented cumulative discounted rewards are:
\begin{equation}
    J = J_1 + \beta*J_2,
\end{equation}
which marks the end of the current episode. Intuitively, the lower the VLM inference performance is, the larger the augmented rewards are. As the RL agent seeks to maximize the cumulative rewards, this leads the agent to perform more actions that push the Unity to generate more challenging images for the VLM to reason the spatial relationships among different objects in each scene. However, the goal of the VLM is to maximize the alignment of image and text modalities such that its spatial reasoning capabilities need to grow. This is, to some extent, similar to the generative adversarial network, differing in the purpose. The former aims at training a decent generator that generates high-quality data, while the latter targets training the VLM (somewhat like a discriminator) to excel at spatial reasoning. A reward scaling factor, $\beta$, is used to balance the reward of these components.

Some recent studies~\cite{yuksekgonul2023when} showcased the potential of utilizing ``hard samples" to fine-tune VLM models. In our work, we developed a similar setup of utilizing ``hard samples" for VLM fine-tuning. For the CLIP-type models, we utilized RL-generated data to curate both ``positive samples" (correct image-caption pair) and ``negative samples" (incorrect image-caption pair) for finetuning VLM, whereas only ``positive samples" are used for PaliGemma. Additionally, it should be noted that due to the model size, we only fine-tune the attention layers of the language model for PaliGemma. Fig.~\ref{fig:spatialCLIP} illustrates the entire fine-tuning pipeline we used, where we feed a batch of data from the RL-generated samples and do a round of inference for feedback to the RL agent each episode $e$. With each episode $e$, we randomly sample image-text pairs with sampling rate $\eta$ to increase sample variety in the collected data for the fine-tuning batch. After $E$ episodes, we fine-tune the VLM with the fine-tuning batch. After fine-tuning is completed, the fine-tuning batch is cleared so that the next batch can be collected for the following iterations. We are now ready to summarize the proposed framework in full.




\begin{algorithm}
\caption{RLS3}
\label{rls3}
\begin{algorithmic}[1]
\State \textbf{Input:} number of iterations $I$, number of episodes per iteration $E$, minimum valid samples per episode $T_0$, sampling rate $\eta$, number of CLIP epochs or PaliGemma steps $K$, VLM reward scaling factor $\beta$
\For{$i=1,2,...,I$}
    \For{$e=1,2,...,E$}
        \State Initialize RL episode
        \For{$t=1,2,...,T \geq T_0$}
            \State Select and execute action $a_t \sim \pi_{\theta}(a|s_t)$ 
            \State Observe reward $r(s_t, a_t)$ and next state $s_{t+1}$
            \State Update SAC agent (policy, value functions)
        \EndFor
        \State Pause on the last step of the episode
        \State Generate prompts using ground truth metadata
        \State Perform VLM inference on batch of size $N = T_0$ samples
        \State Compute VLM reward as $\beta \times r_{VLM}^2$ and return to agent
        \State Randomly sample generated data with sampling rate $\eta$
    \EndFor
    \State Compile fine-tuning batch of size $E \times (\eta \times T_0)$
    \For{$k=1,2,...,K$} 
        \State Fine-tune VLM one epoch (CLIP) or step (PaliGemma)
        \If{$k \mod F = 0$}
            \State Evaluate and log validation metrics
        \EndIf
    \EndFor
    \State Clear fine-tuning batch
\EndFor
\end{algorithmic}
\vspace{-2pt}
\end{algorithm}

\subsection{RL-Based Synthetic Sample Selection}
The RLS3 framework can be seen in full in Fig.~\ref{fig:spatialCLIP} and is described in Algorithm~\ref{rls3}. We generate the synthetic data by utilizing the Unity environment, RL agent, prompt generator, and VLM models. To keep all these components in sync, a scheduler Python script coordinates the execution of the task-specific scripts. The RL process generates data with the Unity environment. First, the RL agent performs an action to place an object at the desired position. Then, the Unity environment captures an image using a camera and generates the ground truth metadata of the objects, which includes the object names, coordinates, and rotation, as well as the camera coordinates and rotation. This process is continued for each step until $T_0 \leq T$ valid images for an episode are generated. The RL script is paused at the end of the last step to wait for an extrinsic VLM-based reward. 
The prompt generator script then takes the ground truth metadata and generates prompts with explicit spatial context. 
When this is done for the entire batch of data, the VLM script takes the image-prompt pairs and does inference to quantify the performance. The inference output is sent to the RL process and is used to generate the extrinsic VLM-based reward, where a low performance corresponds to a high reward. After this reward is applied to the RL agent at the end of the episode, the next episode can begin. In order to increase data diversity, $E$ episodes occur each fine-tuning iteration. The generated data from each episode is sampled with sampling rate $\eta$ and appended to the fine-tuning batch, resulting in a size of $E \eta T_0$. 
When the fine-tuning step is complete, the collected data is cleared, and the next iteration begins.

\section{Experiment Setup}
Our experiments aimed to explore three main categories: random and RL-guided data generation, loop configuration for the RL feedback frequency and amount of generated data per batch, and the spatial reasoning performance of various VLMs on several datasets. 

\subsection{Data Generation Agents}
For RLS3, we train an SAC-based RL agent that receives a reward signal from the VLM. As a baseline, we consider randomly sampled data which would be a standard data generation approach for supervised fine-tuning. We implement the baseline using a random agent that performs random actions without taking into consideration any feedback from the environment or the VLM. To maintain a fair comparison, we test both agents under the same configurations, meaning that the random agent also generates batches of data for fine-tuning the VLM in an episodic manner.

\subsection{Framework Configuration}
Our framework has several parameters that determine the RL feedback frequency, the amount of data used to fine-tune the VLM each iteration, and the length of fine-tuning the VLM itself. These parameters are as follows: number of valid images per RL episode $T_0$, number of sampled valid images per RL episode determined by the sampling rate $\eta$, and number of RL episodes per iteration $E$. The RL feedback frequency is a function of the number of $E$, where a higher number enables more VLM feedback to the agent between iterations fine-tuning the VLM. We use the VLM reward scaling factor $\beta$ to balance the intrinsic feasibility and extrinsic VLM-based rewards. The length of fine-tuning the VLM $K$ is equivalent to the number of epochs for CLIP-type models and the number of steps for PaliGemma. In our experiments, we use the following combination based on extensive hyperparameter search, $T_0=200$, $\eta=0.5$, $E=20$, $\beta=10$, and $K=256$ for PlaiGemma; $T_0=200$, $\eta=0.5$, $E=20$, $\beta=10$, and $K=10$ for CLIP. We additionally stop the loop early if there isn't significant improvement for 10 iterations after at least 15 iterations for PaliGemma, and if there isn't significant improvement for 5 iterations after at least 10 iterations for CLIP. Early stopping is only done when the RL agent is being used, and the resulting total number of generated samples is used as the budget for the runs with a random agent.



\subsection{Datasets}
The training data is generated dynamically from a simulated Unity environment with 5 scenes, allowing for a diverse set of frames that update every episode and fine-tuning batch. By generating new frames each iteration, we ensure that the model is continually exposed to varied scenes and interactions.

For validation, we use the same Unity environment as the training data but create a fixed set of $500$ frames beforehand. This dataset remains static throughout all iterations, providing a consistent benchmark to evaluate the model’s improvements after each training cycle. By holding the validation data constant, we can measure the model’s incremental performance without introducing new variations, helping to isolate the effects of iterative fine-tuning.

The test data is created from a separate Unity environment with 3 unique scenes, designed to evaluate the model’s performance on $1000$ entirely unseen samples. Like the validation set, the test dataset is generated beforehand and remains static, allowing us to assess the model’s robustness in generalizing to different environments. The diversity in scene content between the training and test environments is crucial for evaluating the model’s adaptability to new contexts.
We do not employ any additional test datasets beyond the unique environment test set. A challenge in evaluating spatial reasoning performance is the varying prompt formats and input-output structures. 

\section{Results and Discussion}
We explore several key areas of our framework to validate the update efficiency and data generation efficiency compared to a random agent. Additionally, we do a detailed analysis of the improvement of spatial reasoning capabilities of models before and after fine-tuning with our framework in the context of per-term performance as well as prompt complexity.

  


\subsection{RL-Guided Efficient Data Generation and Model Updates}

\begin{figure*}[t]
  \centering

  \begin{minipage}{0.925\textwidth}
    \centering
  \begin{subfigure}[b]{0.48\textwidth}
    \centering
    \includegraphics[width=\linewidth]{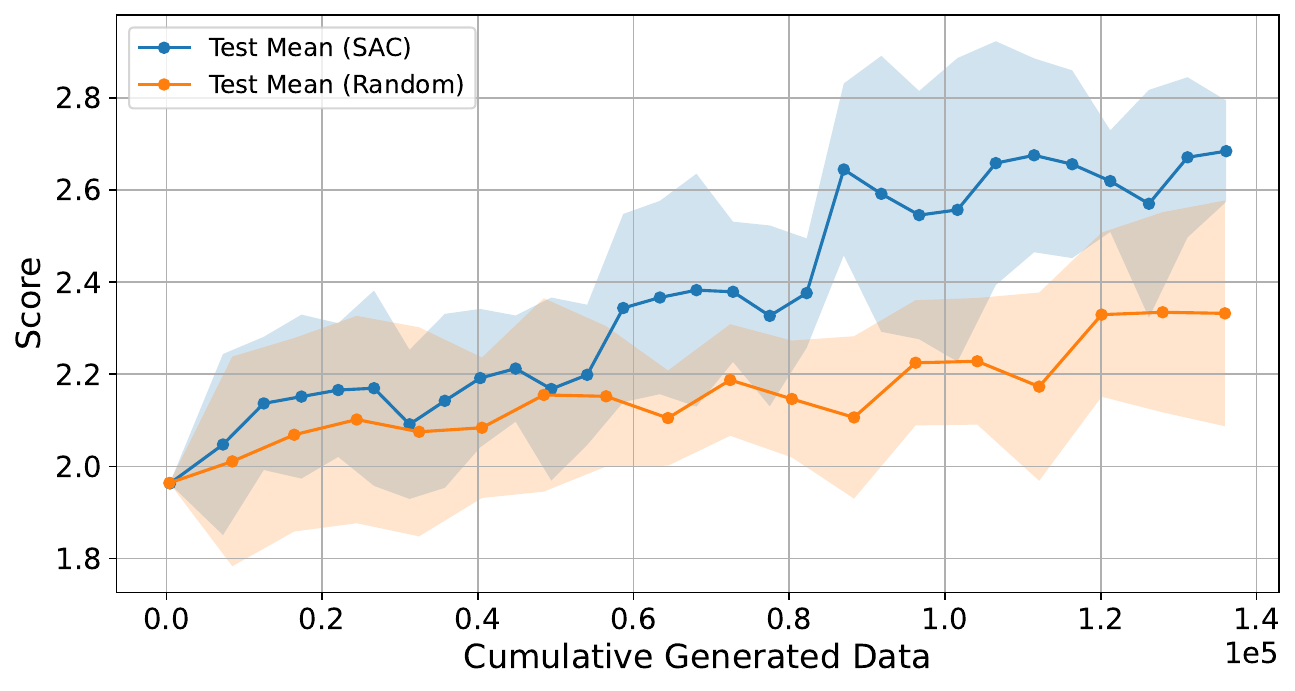}
    \caption{PaliGemma score (avg $\pm std$) on testing data vs cumulative generated data across 5 runs for both SAC and random agents.}
    \Description{A plot of PaliGemma score on testing data on the y-axis and the cumulative generated data on the x-axis. Plots for both SAC and random agents are shown, with the SAC agent having better performance. \textcolor{blue}{Update when last random run finishes}}
    \label{fig:pg_data}
  \end{subfigure}
  \hfill
  \begin{subfigure}[b]{0.48\textwidth}
    \centering
    \includegraphics[width=\linewidth]{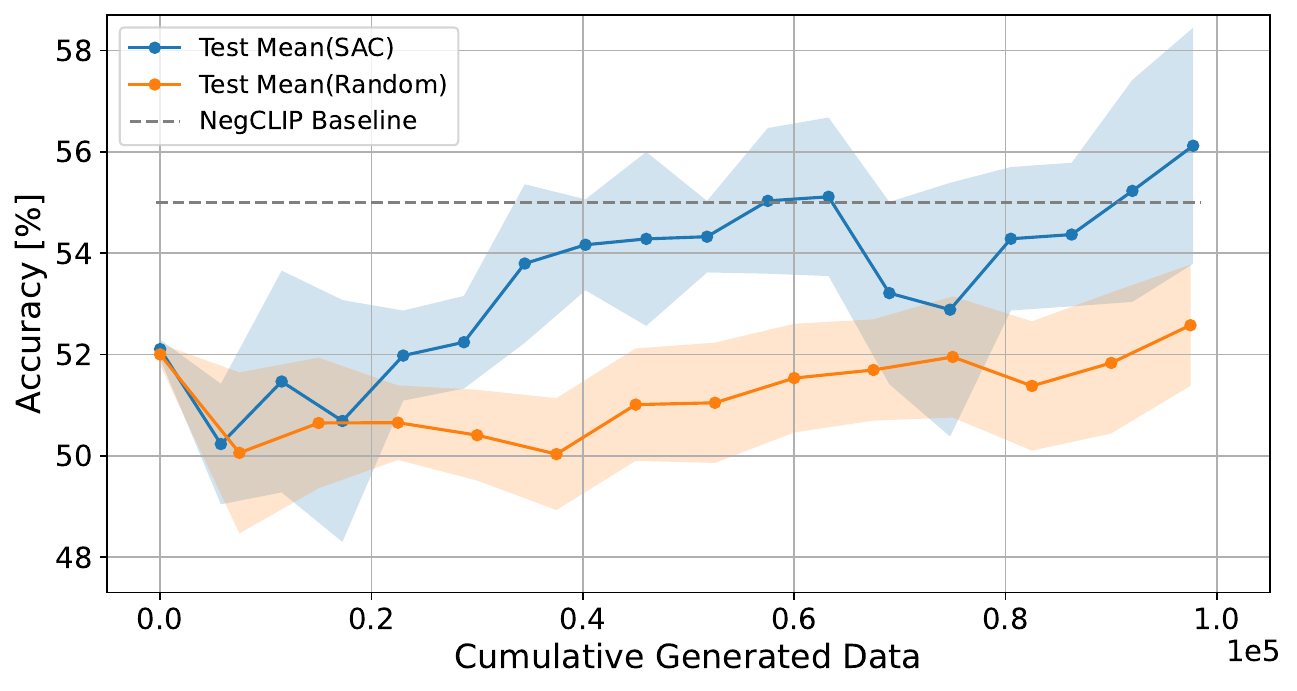} 
    \caption{CLIP accuracy (avg $\pm std$) on testing data vs cumulative generated data across 3 runs for both SAC and random agents.}
    \Description{Plots showing the relative average increase in spatial reasoning performance for both PaliGemma and CLIP. }
    \label{fig:clip_data}
  \end{subfigure}
  \end{minipage}

  \caption{Combined figures of PaliGemma score and CLIP performance vs cumulative generated data.}
  \label{fig:data_efficiency}
\end{figure*}

The integration of RL into RLS3 enhances sample generation efficiency, as seen in Fig. \ref{fig:data_efficiency}. By learning and adapting to the dynamics of the environment, the RL agent is able to maximize the yield of valid samples, thereby reducing the amount of data generated overall. This efficiency stems from the RL agent's capacity to differentiate between promising and unpromising actions, producing more relevant and useful samples with fewer discarded, invalid instances. Even though the environment used in this study is relatively inexpensive, it is evident that this strategic sampling approach is particularly beneficial for high-cost environments where data generation is resource-intensive or time-consuming.


Due to the high computational expense of fine-tuning large models such as VLMs, maximizing the information gained from each sample becomes crucial to maintaining efficiency. Each point in Fig. \ref{fig:data_efficiency} corresponds to a fine-tuning iteration. The random agent configuration experiences approximately 40\% fewer iterations due to the greater number of invalid samples generated. 
RLS3 shows consistently higher performance with an initially widening gap between the SAC and random agents for both PaliGemma and CLIP. This advantage suggests that RL-driven data generation strategies can prioritize more informative samples and allow for a sharper increase in model performance with fewer iterations. 
As the number of iterations grows, RLS3 maintains its advantage but experiences a stagnation in performance and is stopped early. 
In contrast, the random generation method displays a steadier but slower performance increase. Furthermore, CLIP with RLS3 is able to achieve comparable performance to NegCLIP. 
This highlights the primary benefit of RLS3 is the more efficient path to performance gains in the early stages of fine-tuning.

\subsection{Spatial Reasoning Performance}
\begin{figure}[h!]
  \centering
    \includegraphics[width=0.8\linewidth]{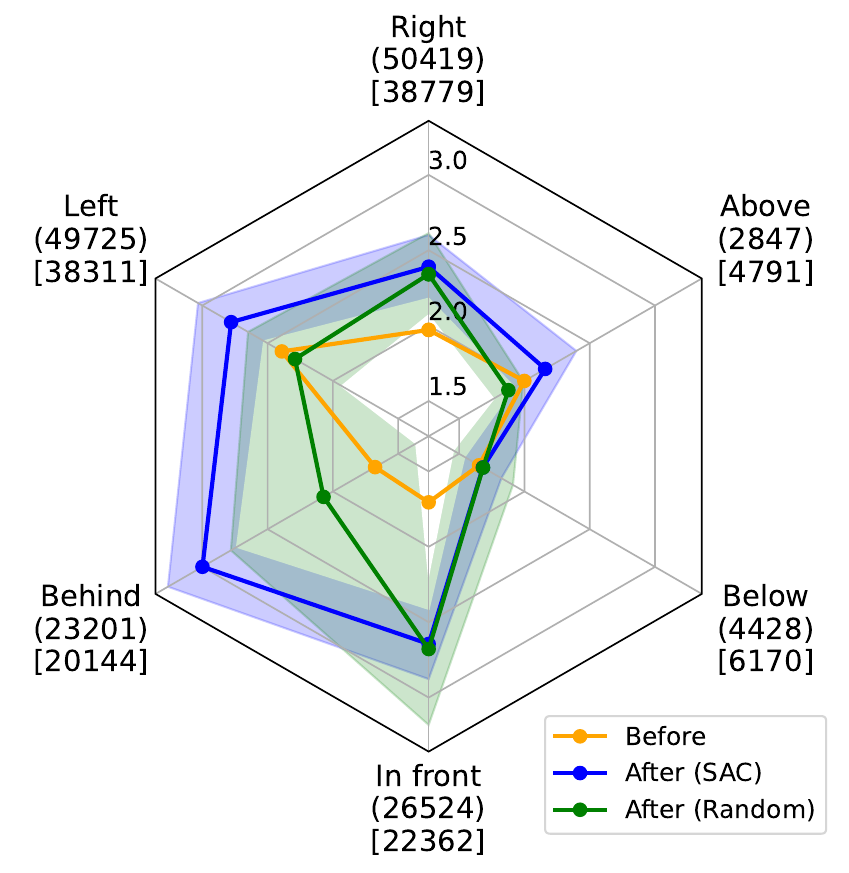}
    \caption{PaliGemma score ($avg \pm std$) separated by spatial term for RLS3 with an SAC and random agent. Cumulative term counts of data generated for fine-tuning are given in ( ) for SAC and [ ] for random agents.}
    \Description{A spider plot showing the improvement of PaliGemma with RLS3 using a random agent and an SAC agent compared to the initial performance. The random agent leads to mixed performance gains while some terms (left and above) decrease. The SAC agent achieves equivalent or better performance for all terms.}
    \label{fig:pg_score_bars}
\end{figure}

The nature of spatial descriptions inherently contains contradictory terms, such as `left' and `right,' which often leads to improvement in one term paired with the decline of the other. Thus, it is important to explore the per-term performance to better quantify the spatial reasoning performance. We explore the `before' and `after' performance for PaliGemma due to the larger relative improvement, as seen in Fig.~\ref{fig:pg_score_bars}. Through our fine-tuning process with an SAC agent, we were able to achieve equivalent or better performance for all spatial terms explored. In contrast, when a random agent is used, `left' and `above' experience a decrease in performance. Furthermore, the SAC agent is able to achieve equivalent or better performance for all terms other than `in front', where the random agent holds a slight advantage. The cumulative number of generated samples containing each spatial term reveals some interesting insights. Most notably, the SAC agent has a much less even distribution of generated samples, heavily preferring samples with either `left' or `right'. The bias towards generating samples of these opposing terms may be partially responsible for the performance of both increasing rather than the model experiencing a tradeoff. Additionally, despite generating fewer samples with `above' or `below,' the SAC agent achieves equivalent performance for `below' and notably better performance for `above.' This highlights that the SAC agent is able to generate informative samples that lead to more sample-efficient learning, which is also highlighted in the `behind' case where the SAC agent achieves notably higher performance with a roughly equivalent number of samples to the random agent. 

Lastly, it should be noted that while the random agent produces prompts with a generally more balanced distribution of spatial terms, there is still a noticeable bias towards horizontal terms, especially `left' and `right.' This is likely caused by inherent biases due to the Unity environment setup. Generally, the surfaces that allow vertical relations to occur have a smaller area than the primary surface, e.g. a cabinet and a countertop. This creates a bias towards horizontal relations. Furthermore, the landscape orientation of the cameras creates a bias towards generating `left' and `right' scenarios. However, these inherent biases also highlight the importance of intelligent sample selection to account for underrepresentation, whether through simply increasing the number of samples or prioritizing informative samples.

\subsection{Effect of Prompt-Complexity}
\begin{figure}[h]
  \centering
  \includegraphics[width=0.9\linewidth]{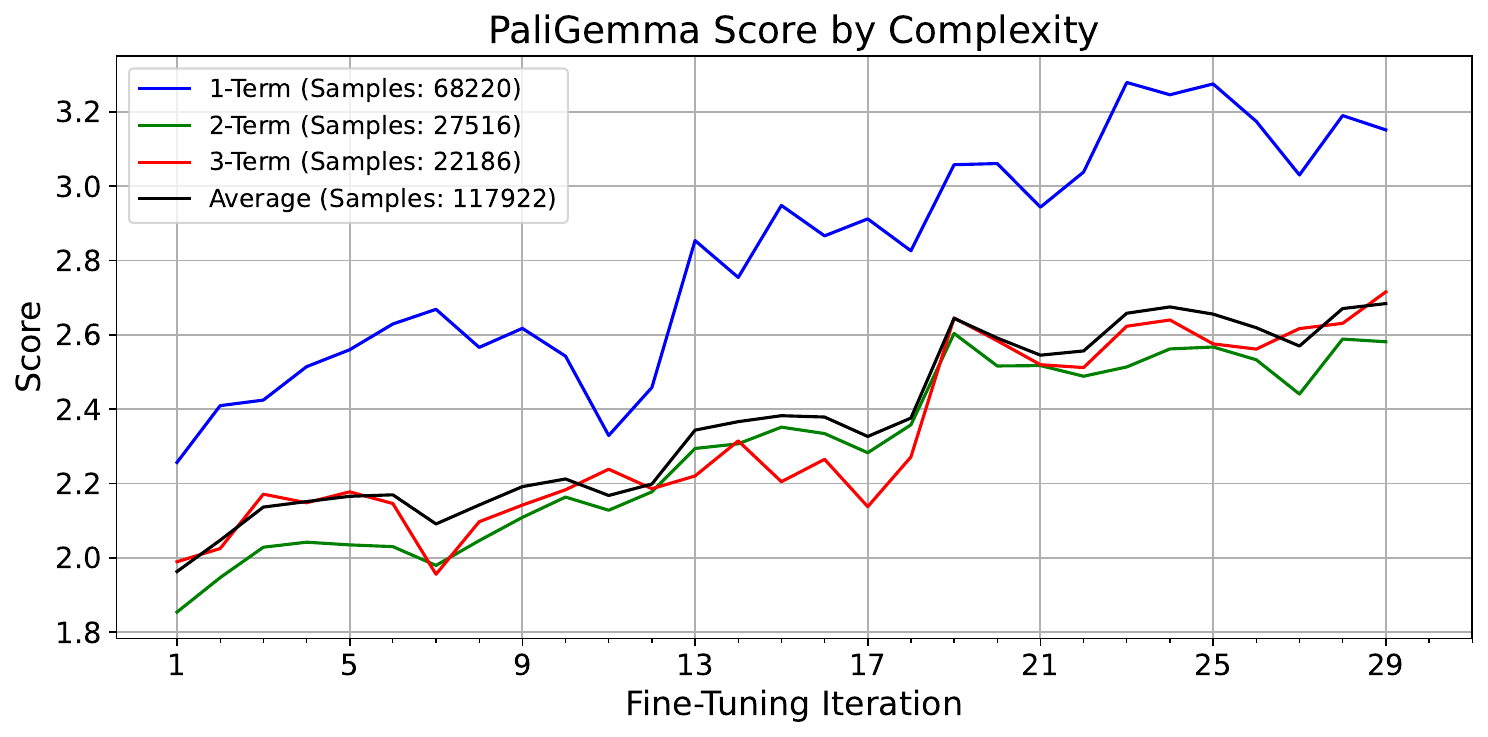}
  \caption{Average PaliGemma score by prompt complexity vs iteration for SAC agent.}
  \Description{A plot showing the average PaliGemma score separated by the number of spatial terms in the testing prompt. Performance is highest for 1-term prompts while 2- and 3-term prompts show lower, but similar performance.}
  \label{fig:pg_term_complexity}
\end{figure}

As discussed in Section~\ref{sec:method}, we define prompt complexity as the number of spatial terms included in each prompt. As seen in Fig.~\ref{fig:pg_term_complexity}, prompts with only a single term are easier for PaliGemma to handle than those with multiple terms, leading to higher performance. This may also be due to the larger number of samples generated with a single term. Interestingly, there does not appear to be a significant difference in the performance of PaliGemma on prompts with 2 and 3 spatial terms. This may be caused by the roughly equivalent number of samples generated for each, along with the rubric used, making scores for partially correct answers similar. 


\subsection{Dynamics of VLM Fine-Tuning}
The structure of the RLS3 framework requires the VLM models to be fine-tuned sequentially across distinct iterations. In each iteration, the VLM is fine-tuned on newly generated training data while preserving the precious iteration's model weights as the starting point. This approach leverages the learning acquired in earlier iterations to further refine the models with each new batch of data. 

\begin{figure}[h]
  \centering
  \includegraphics[width=0.9\linewidth]{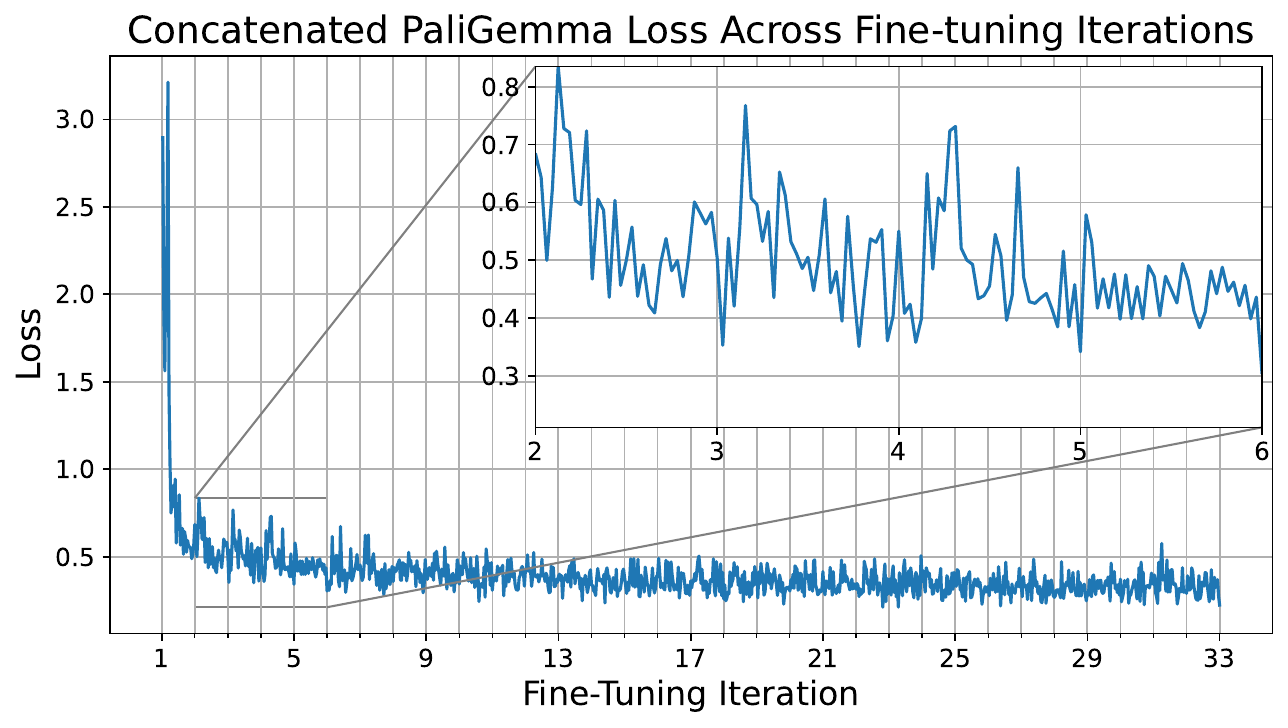}
  \caption{Concatenated PaliGemma loss plots for iterative fine-tuning.}
  \Description{A plot showing a large initial decrease in loss. The x-axis indicates the iterations with shows a cyclic pattern in the concatenated losses where a spike occurs at the start of each iteration. These spikes decrease as the number of iterations increases, leading to convergence.}
  \label{fig:pg_loss}
\end{figure}

As illustrated in Fig.~\ref{fig:pg_loss}, this iterative fine-tuning process creates a recurring pattern of loss spikes at the start of each iteration. These spikes are attributed to the VLM adjusting to the novel samples introduced, which may differ in complexity or characteristics from the prior data. While the loss spikes are more pronounced in the earlier iteration, they tend to decrease progressively as the model adapts to the distribution of the data over time. This trend suggests that the VLM is retaining useful information from prior iterations and stabilizing as it generalizes across a wider variety of samples. 

\begin{figure}[h!]
    \centering
    \includegraphics[width=0.9\linewidth]{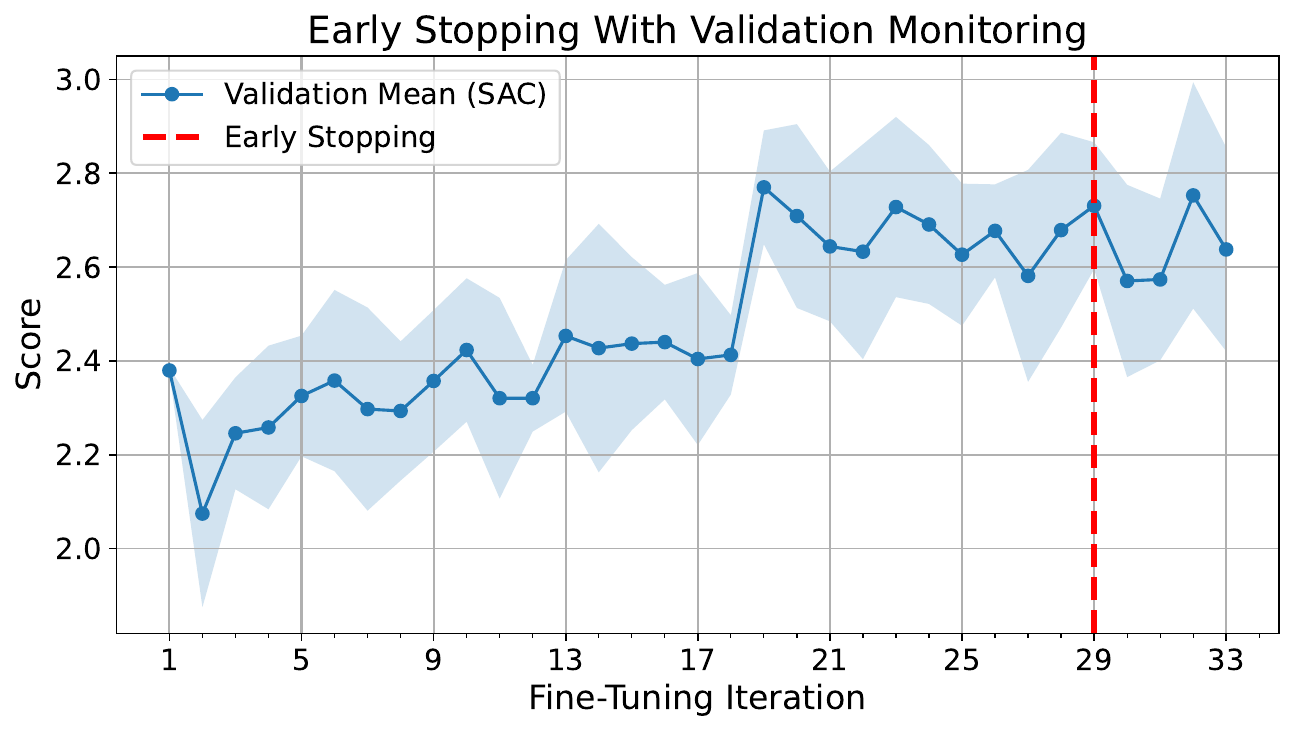}
    \caption{PaliGemma score (avg $\pm std$) on validation data vs fine-tuning iteration across 5 runs for the SAC agent with the early stopping point indicated.}
    \Description{A plot of PaliGemma score on validation data on the y-axis and the cumulative generated data on the x-axis. Plot for SAC agent with early stopping indicated at iteration 29.}
    \label{fig:pg_early_stop}
  \end{figure}

To avoid running RLS3 longer than necessary, we utilize early stopping criteria, as discussed in Section~\ref{sec:method}, which is facilitated through monitoring the validation performance as shown in Fig.~\ref{fig:pg_early_stop}. The observed trend in the validation performance mirrors that of testing, suggesting that the criteria are appropriately chosen.

\section{Conclusions and Future Work}
In this work, we introduced RLS3, a novel framework that integrates RL with a VLM to iteratively improve spatial reasoning performance. RLS3 effectively utilizes an RL agent to generate feasible and informative samples, guided by feedback from both the environment and the VLM. This approach results in greater initial performance gains in spatial reasoning as the agent explores relevant regions of the state-action space. By continuously fine-tuning the VLM with generated data, we achieve more targeted fine-tuning.

Future work aims to expand the versatility and scalability of the RLS3 framework. One key direction is to adjust the framework for multi-device communication, enabling more efficient distributed training and allowing for fine-tuning of larger models. This extension would support more sophisticated models, such as LLaVA and Grounding DINO, further enhancing the framework's ability to handle complex multimodal tasks. Additionally, we plan to explore applications where synthetic data generation is expensive or constrained, where the data efficiency of RLS3 could minimize resource use without compromising sample quality. However, another important consideration when using synthetic data is the challenge of sim2real transfer which in itself necessitates further investigation, perhaps through the direct integration of domain adaptation techniques. Another direction to explore is the mechanism by which the RL agent receives feedback from the VLM. Rather than batch-wise feedback, finer-grained signals would make the reward less sparse and more precisely identified informative samples.

\begin{acks}
This work was partly supported by the National Science Foundation, USA, under grants CAREER CNS-1845969 and CPS Frontier CNS-1954556.
\end{acks}

\bibliographystyle{ACM-Reference-Format}
\bibliography{references}

\end{document}